\documentclass[sigconf]{acmart}
\AtBeginDocument{%
  }

\copyrightyear{2026}
\acmYear{2026}
\setcopyright{cc}
\setcctype{by}
\acmConference[WWW '26]{Proceedings of the ACM Web Conference 2026}{April 13--17, 2026}{Dubai, United Arab Emirates}
\acmBooktitle{Proceedings of the ACM Web Conference 2026 (WWW '26), April 13--17, 2026, Dubai, United Arab Emirates}
\acmPrice{}
\acmDOI{10.1145/3774904.3793042}
\acmISBN{979-8-4007-2307-0/2026/04}

\usepackage[utf8]{inputenc} 
\usepackage[T1]{fontenc}    
\usepackage{hyperref}       
\usepackage{url}            
\usepackage{booktabs}       
\usepackage{amsfonts}       
\usepackage{nicefrac}       
\usepackage{microtype}      
\usepackage{xcolor}         
\usepackage{wrapfig}
\usepackage{graphicx}
\usepackage{subfigure}
\usepackage{multicol}
\usepackage{multirow}
\usepackage[ruled,linesnumbered]{algorithm2e}
\usepackage{subcaption}
\usepackage{amsmath}

\begin{document}

\title{SEP-Attack: A Simple and Effective Paradigm for Transfer-Based Textual Adversarial Attack}



\author{Han Liu}
\affiliation{
  \institution{Dalian University of Technology}
  \city{Dalian}
  \country{China}}
\email{liu.han.dut@gmail.com}

\author{Zhi Xu}
\affiliation{
  \institution{Dalian University of Technology}
  \city{Dalian}
  \country{China}}
\email{xu.zhi.dut@gmail.com}

\author{Xiaotong Zhang}
\authornote{Corresponding author.}
\affiliation{
  \institution{Dalian University of Technology}
  \city{Dalian}
  \country{China}}
\email{zxt.dut@hotmail.com}

\author{Feng Zhang}
\affiliation{
  \institution{Peking University}
  \city{Beijing}
  \country{China}}
\email{fengzhangyvonne@gmail.com}

\author{Xiaoming Xu}
\affiliation{
  \institution{Dalian University of Technology}
  \city{Dalian}
  \country{China}}
\email{xu.xm.dut@gmail.com}

\author{Wei Wang}
\affiliation{
  \institution{Macao Polytechnic University}
  \city{Macao}
  \country{China}}
\email{weiwang@mpu.edu.mo}

\author{Fenglong Ma}
\affiliation{
  \institution{The Pennsylvania State University}
  \city{Pennsylvania}
  \country{United States}}
\email{fenglong@psu.edu}

\author{Hong Yu}
\affiliation{
  \institution{Dalian University of Technology}
  \city{Dalian}
  \country{China}}
\email{hongyu@dlut.edu.cn}

\renewcommand{\shortauthors}{Han Liu et al.}

\begin{abstract}
Despite the strong performance of deep neural networks in modern Web and language applications, they remain vulnerable to adversarial attacks, especially transferable attacks that generate adversarial examples using surrogate models without accessing the victim model. Transferable attacks in the text domain are still under-explored, with only a few studies addressing this challenging issue, often with suboptimal results due to equal treatment of submodels or inaccurate estimation of importance scores. To address these challenges, we propose a simple yet effective paradigm for transfer-based textual adversarial attack, named SEP-Attack. Specifically, we employ the Determinantal Point Process (DPP) to generate diverse surrogate ensemble weights, representing the transferability of submodels. Using these weights, we introduce a new metric to evaluate prediction confidence scores, which in turn are used to calculate word importance scores and generate adversarial candidates. Finally, we quantify the transferability score for each candidate and select the top ones as the final transferable adversarial examples. Experiments conducted on four datasets and two real-world APIs validate the efficacy of SEP-Attack, significantly outperforming state-of-the-art baselines.
\end{abstract}



\maketitle

\section{Introduction}

Deep neural networks (DNNs) and large language models (LLMs) have become the backbone of modern Web-based applications, supporting tasks such as content moderation~\cite{cm1,cm2,cm3}, online recommendation~\cite{onl1,onl2,onl3}, and misinformation detection~\cite{misinfo1,misinfo2,misinfo3}. Despite their widespread adoption and strong performance, growing evidence shows that these models remain highly vulnerable to adversarial attacks~\cite{why-imperceptible,stealthy-porn}. Such vulnerabilities raise serious societal and ethical concerns when models are deployed on the Web, where adversarial attacks can amplify harmful content, evade moderation systems, or trigger biased or misleading decisions. Understanding such risks is vital for building safe and reliable Web AI systems.

\begin{figure*}[t]
    \centering
    \includegraphics[width=0.8\linewidth]{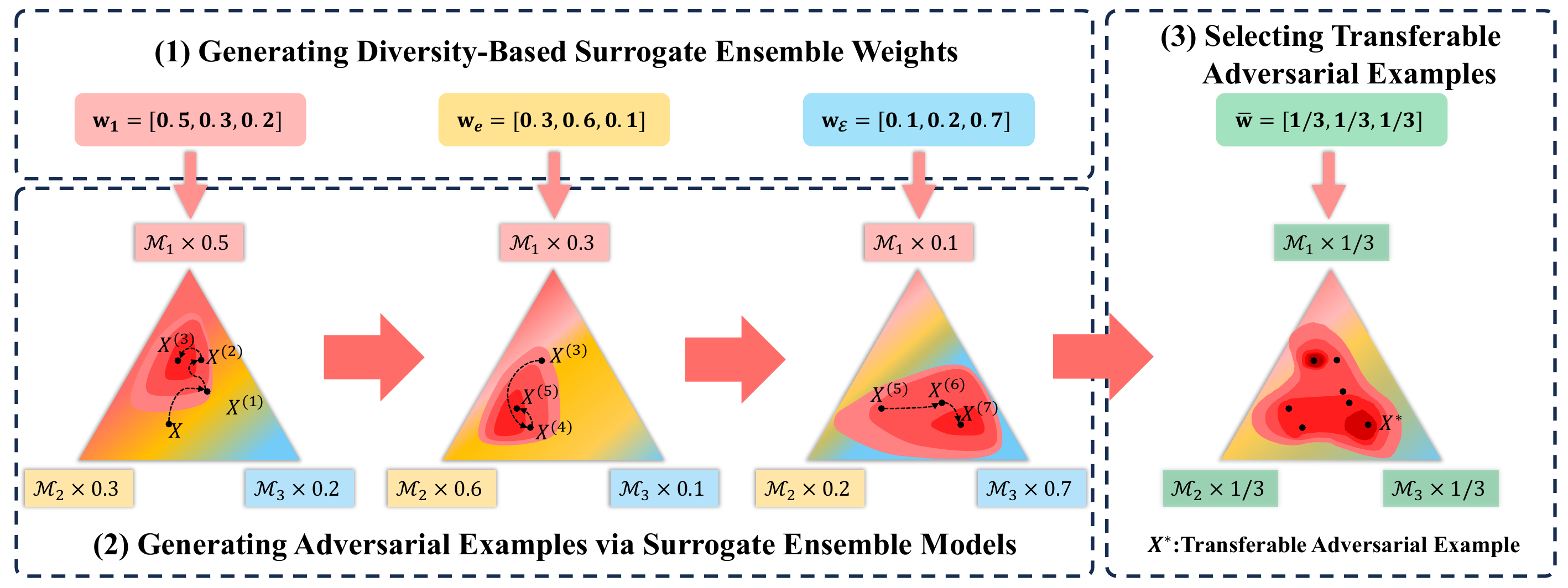}
    \caption{The overview of our proposed SEP-Attack framework, which consists of three main components.
(1) Generating diversity-based surrogate ensemble weights with the Determinantal Point Process (DPP). (2)
Generating adversarial examples via the surrogate ensemble model, with deeper red hues representing lower confidence scores under corresponding weights; and (3) Selecting transferable adversarial examples according transferability scores.
}
    \label{fig:frame}
\end{figure*}

Existing adversarial attack methods can be divided into two categories based on the accessibility level of victim models: \emph{white-box attacks}~\cite{hotflip,whitebox2} and \emph{black-box attacks}~\cite{bertattack,blattack2}. 
\emph{White-box attacks} allow the attacker full access to the victim model, including its architecture, parameters, and training data. Although these attacks are highly idealistic and impractical in most real-world scenarios, they are still valuable in controlled environments such as internal testing and model sharing. In contrast, \emph{black-box attacks} assume that the attacker has no knowledge of the model's internal workings, which is more realistic in practical applications. Black-box attack methods can be further classified into \emph{query-based attacks} and \emph{transfer-based attacks}. \emph{Query-based attacks} need to frequently query the victim model until an adversarial example is generated, which are probably limited by query restrictions and detection mechanisms. \emph{Transfer-based attacks} aim to generalize adversarial vulnerabilities across different models without accessing any information of the victim models, making them particularly concerning for real-world applications. Most existing transfer-based attacks have primarily focused on computer vision~\cite{rap,2024HuanranChengRethinking,2017YanpeiLiu,pgn}, where they often achieve their attacks by setting a target function and utilizing gradient descent on a surrogate model. However, these approaches face challenges in text attacks due to the discrete nature of text and variations in tokenization algorithms across different models~\cite{yuan2021transferability,ensemble}. This challenging issue is still in its infancy and under-explored in the text domain.

HAWR~\cite{yuan2021transferability} proposes a method to generate adversarial examples by deriving highly-transferable adversarial word replacement rules through analyzing the changes in the ensemble's log-likelihood predictions caused by word substitutions. Ensemble~\cite{ensemble} generates a transfer-based adversarial example by leveraging TextFooler~\cite{Textfooler} to generate adversarial examples on an ensemble model and subsequently transfer them to the victim model for the attack. However, these methods have several limitations when it comes to generating transferable text adversarial examples.
On the other hand, the TextFooler-based approach calculates word importance by removing words from the sentence. This removal alters the context, causing inaccurate importance estimates that compromise the attack success rate.

To address these challenges and to support robustness research for responsible Web AI, we propose SEP-Attack, a \textbf{S}imple yet \textbf{E}ffective \textbf{P}aradigm for transfer-based textual adversarial \textbf{Attack}. Specifically, "simple" refers to the straightforward and easily implementable core concept of SEP-Attack. Meanwhile, "effective" indicates that SEP-Attack demonstrates superior performance, significantly outperforming other strong baselines. An overview of SEP-Attack is illustrated in Figure~\ref{fig:frame}. In SEP-Attack, we propose to utilize the Determinantal Point Process (DPP) to sample diverse surrogate ensemble weights to represent the levels of the transferability of submodels, resulting in a set of transferable adversarial examples. To address the challenge of accurately calculating word importance, we propose an approach that involves updating the important words and subsequently removing the unimportant ones. This strategy aims to mitigate the issues arising from inaccurate word importance calculations. Finally, we identify transferable adversarial examples among all the examples generated from the ensemble model. We conduct experiments on four benchmark datasets, two real-world APIs (Alibaba Cloud and Google Cloud), and large language models. The results demonstrate that SEP-Attack achieves a significantly higher attack success rate compared to other strong baselines while also maintaining the lowest query budget cost\footnote{The source code is publicly available at https://github.com/ZevineXu/SEP-Attack}.

\section{Related Work}

\subsection{Query-Based Black-Box Attacks}
Query-based black-box attacks aim to generate adversarial samples by querying the black-box victim models. Two main ways to achieve query-based attacks are the soft and hard-label settings. In the soft-label setting, attackers can access the model's output scores (i.e., probability distributions) and use them to design attack algorithms~\cite{DeepWordBug, Textfooler, TextBugger, BBA}. For example, DeepWordBug~\cite{DeepWordBug} focuses on generating character-level perturbations on important words using output scores.  TextFooler~\cite{Textfooler} uses output scores to calculate word-level importance and performs attacks by a greedy-based method. TextBugger~\cite{TextBugger} aims to identify important sentences and replace words within these sentences to achieve attack objectives. The scores can also be used to train a surrogate model to achieve the attacks, such as BBA~\cite{BBA}. 
Unlike the soft-label scheme, the \textit{hard-label} setting only allows attackers to obtain the model's predicted labels~\cite{HLGA, ye2023pat, LeapAttack, TextHoaxer, hqa}. For instance, LeapAttack~\cite{LeapAttack} estimates the position of the decision boundary using a Monte Carlo estimation method and improves the success rate of hard-label adversarial attacks by gradually moving examples closer to the decision boundary. HQA-Attack~\cite{hqa}, the state-of-the-art model, enhances the success rate of adversarial attacks by finding the transition word, estimating the direction of word vectors and continuously improving the quality of examples. Although query-based approaches have achieved high success rate, they still require a large number of queries to the victim model, which is difficult to achieve in real-world applications.

\subsection{Transfer-Based Black-Box Attacks}
Transfer-based black-box attacks aim to generate adversarial examples by manipulating gradients~\cite{Mi_FGSM,e_mi_FGSM} or leveraging surrogate ensembles~\cite{pgn,Mi_FGSM,yuan2021transferability,ensemble} to attack victim models. However, most existing work focuses on image attacks~\cite{Mi_FGSM,e_mi_FGSM,Mi_FGSM}, which are not directly applicable to the text domain due to its discrete nature and tokenization variations. 
Only a few models have been proposed recently in the text domain, which are all ensemble-based transfer attacks. Existing work HAWR~\cite{yuan2021transferability} generates highly-transferable adversarial examples by identifying and applying word replacement rules that significantly alter the predictions of an ensemble of neural text classifiers. These rules are derived by analyzing the changes in log-likelihood caused by word substitutions in the ensemble's predictions, allowing the creation of adversarial examples that can effectively fool multiple models. However, this work ignores semantic differences between different texts, thus leading to a low success rate of attacks for the generated adversarial examples. 
Another work Ensemble~\cite{ensemble} generates adversarial examples on the ensemble model based on TextFooler and then performs transfer attacks. This method addresses the issue of conflicts between different tokenization algorithms by employing synonym replacement. However, TextFooler calculates word importance by removing words without considering the problem of semantic changes in the text, resulting in the inaccurate calculation of word importance. Moreover, the TextFooler algorithm can only iterate once on the adversarial examples, which hampers its ability to effectively optimize the generated adversarial examples. Consequently, the success rate of this ensemble attack method is also low. 

\section{The Proposed Paradigm}

\subsection{Problem Formulation}
In this work, we focus on the transfer-based adversarial attack which use the surrogate ensemble model $\mathcal{M} = \{\mathcal{M}_1, \cdots, \mathcal{M}_N\}$, and $\mathcal{M}_n$ is the $n$-th submodel.
Given a text $X=[x_1, \cdots, x_i,\cdots,x_L]$ with the ground truth label $y$, where $x_i$ represents the $i$-th word and $L$ denotes the length of $X$, the goal of this work is to construct a set of potential adversarial examples $\mathcal{X}^\prime = \{X'_1, \cdots, X'_K\}$ using the ensemble model $\mathcal{M}$, where $K$ is the number of selected adversarial examples. Each potential adversarial example $X'_t = [x_1', \cdots ,x_i', \cdots ,x_L']$ is generated by replacing the original word $x_i$ where part-of-speech (POS) is \textit{noun}, \textit{verb}, \textit{adverb}, \textit{adjective} with a synonym $x_i'$ in the synonym set $\mathcal{S}(x_i)$. An attack succeeds if any sample $X'_k \in \mathcal{X}^\prime$ leads the victim model $f$ to a wrong prediction, i.e.,
\begin{equation}
    \label{eq:definition}
    f(X'_k) \neq f(X), \ \   \mbox{s.t.} \ \   d(X'_k,X) \leq \epsilon, \exists k \in [1, K],
\end{equation}
where $d(\cdot,\cdot)$ is a distance metric that measures the level of perturbation between two texts, and $\epsilon$ is the predefined hyperparameter. Next, we provide the details of constructing potential adversarial examples $\mathcal{X}^\prime = \{X'_1, \cdots, X'_K\}$ using the ensemble model $\mathcal{M} = \{\mathcal{M}_1, \cdots, \mathcal{M}_N\}$.

\subsection{Generating Diversity-Based Surrogate Ensemble Weights}
Intuitively, the ability to generate adversarial examples for each submodel in $\mathcal{M}$ is different. Thus, equally treating them when generating transferable adversarial examples is not reasonable. To enhance the attack success rate, an ideal solution is to learn a set of weights for submodels. However, this is challenging since we cannot query the victim model to obtain any feedback as guidance in optimizing the weights.

To solve this issue, we propose to employ the \textbf{Determinantal Point Process} (DPP)~\cite{dpp} to estimate the best subset of weights in the whole weight space. DPP is commonly employed in recommendation systems to sample diverse subsets to generate a set of weights. The aim of diversification is to increase the discrepancy among adversarial examples. Specifically, we first randomly generate a weight space $\mathcal{W} \in \mathbb{R} ^{D \times N}$, where $\mathcal{W}_{dn} \in [0,1]$, $D$ denotes the size of the space, and each row is a random weight vector for the ensemble model.

We then calculate the weight kernel matrix $\mathcal{K} = \mathcal{W} \times \mathcal{W}^\top$. Subsequently, the DPP algorithm samples a diverse subset based on the following probability formula:
\begin{equation}
    P(\mathcal{E}) = \frac{\det(\mathcal{K}_\mathcal{E})}{\det(\mathcal{K})},
\end{equation}
where $\mathcal{K}_\mathcal{E}$ is the submatrix of $\mathcal{K}$ corresponding to the selected subset $\mathcal{E}$, and $P(\mathcal{E})$ is the probability of selecting subset $\mathcal{E}$. Intuitively, the largest $P(\mathcal{E})$ indicates the subset $\mathcal{E}$ is the most representative, containing sufficient yet diverse information to represent the entire weight space. Thus, the output of the DPP is the estimated diversity-based weights $\mathcal{W}_\mathcal{E} = [\mathbf{w}_1, \cdots, \mathbf{w}_E]$ for the ensemble model, where $E$ is the size of the best representative subset $\mathcal{E}$.

\subsection{Generating Adversarial Examples via the Surrogate Ensemble Model}
\label{sec:generation}

In the image domain, perturbations can be easily optimized by gradient-based methods, such as FGSM~\cite{fgsm} and PGD~\cite{PGD}. However, these approaches are not easily applicable to text world due to the discrete nature and variations in tokenization algorithms. In the text domain, TextFooler~\cite{Textfooler} tries to overcome this by calculating word importance and sequentially replacing important words to achieve the desired attack. However, accurately determining word importance is difficult because the text undergoes constant changes during the replacement process. To tackle these issues, our proposed model SEP-Attack repeats the following three steps $T$ times: (1) Determining the replacement order, (2) Updating potential important words, and (3) Removing less important words. Before formally introducing the three steps, we first define the importance score of a text. The pseudocode is shown in Algorithm~\ref{alg:generate}.

\begin{algorithm}[t]
\caption{Generate Adversarial Examples via the Surrogate Ensemble Model}
\label{alg:generate}
\SetKwInOut{Input}{Input}\SetKwInOut{Output}{Output}\SetKw{}{}
\Input{The original text $X=[x_1,...,x_L]$,the ground truth label $y$, the threshold $\eta$, the threshold $\epsilon$, the surrogate ensemble weight $\mathcal{W}_{\mathcal{E}}$, the distance function $d(\cdot,\cdot)$.}
\Output{The potential adversarial examples $\mathcal{X}$}
\BlankLine
$X'_0 \leftarrow X$ \\
\For{$\mathbf{w}_e \, \mathbf{in} \, \mathcal{W}_{\mathcal{E}}$}{
    $t \leftarrow 0$\\
    \While{$t \leq T$} {
        \For{$x_i \, \mathbf{in} \, X$} { 
            $I_{x_i} = -\mathcal{\delta}(X_{\textbackslash x_i},y;\mathbf{w_e} )$ \\
            $\mathbf{I}$.insert($I_{x_i}$) \\
        }
        Sampling the updating order $\mathbf{O}$ according to Eq. (\ref{eq:sampling_distribution}) \\
        \For{$x_i \, \mathbf{in} \, \mathbf{O}$}{
               Update $x_i'$ in $X'_t$ with $x^*_i$ by Eq. (\ref{eq:update_word}) \\
               \If{$d(X'_t,X) \geq \eta$}{
                    \textbf{break;} \\
               }
        }
        \While{$d(X'_t,X) \geq \epsilon$} {
            $\mathbf{C} \leftarrow $ Get the positions of all different words between $X$ and $X'_t$ \\
            \For{$i \, \mathbf{in} \, \mathbf{C}$}{
                $\alpha_i $= $\mathcal{\delta}(X'_{[x_i' \rightarrow x_i]},y;\mathbf w_e)$ \\
                $choices$.insert($\alpha_i$); \\
            }
            Find the minimum $\alpha_i^*$ and $X'_{[x_i'^* \rightarrow x_i^*]}$ from $chocies$\\
            Update $x_i'$ in $X'_{t+1}$ with $x_i^*$\\
        }
        $\mathcal{X}$.insert($X'_{t+1}$)\\
        $t=t+1$\\
    }
}
\textbf{return} $\mathcal{X}$
\end{algorithm}

\subsubsection{Confidence Score Calculation}
For each weight vector $\mathbf{w}_e = [w_1^e,\cdots, w_N^e] \in \mathcal{W}_{\mathcal{E}}$, we can evaluate the confidence score of a prediction for a given text $\bar{X}$ in a weighted manner as follows:
\begin{equation}
    \label{eq:loss}
    \delta(\bar{X}, y, \mathbf{w}_e) = \sum_{n=1}^{N} w_n^e * \Bigl(\mathcal{M}_n(\bar{X}, y) - \mathop{\max}\limits_{y'\neq y}\mathcal{M}_n(\bar{X},y')\Bigr),
\end{equation}
where $\mathcal{M}_n(\bar{X},y)$ denotes the probability score assigned to the ground truth label $y$, and $\mathcal{M}_n$ is the $n$-th submodel. A higher $\delta$ value indicates a  confident prediction for data $\bar{X}$ on label $y$. In the context of transfer-based adversarial attacks, our objective is to find a example $\bar{X}$ that minimizes the confidence score $\delta(\bar{X}, y, \mathbf{w}_e)$. By reducing $\delta$, we aim to decrease the model's confidence in its prediction, thereby increasing the likelihood of a successful attack.

\subsubsection{Determining the Replacement Order}

This step aims to determine a suitable replacement order. We use a sampling method to determine the replacement order to ensure the diversity of the generated adversarial examples in the ensemble model. Specifically, we can represent the sentence $X$ after removing the word $x_i$ as $X_{\textbackslash x_i} = [x_1,\cdots,x_{i-1},x_{i+1},\cdots, x_L]$. In this context, the importance score of $x_i$ can be defined as follows:
\begin{equation}\label{eq:word_importance}
    I_{x_i} = -\delta(X_{\textbackslash x_i},y, \mathbf{w}_e).
\end{equation}
If a word $x_i$ is important, it will decrease the confidence level of the prediction and vice versa. Thus, the importance of a word is negatively related to the prediction confidence.

Since we tend to prioritize the replacement of more important words, it is important to consider that the differences between $I_x$ values may not be substantial, especially in long text. Consequently, relying solely on softmax may not effectively bias the sampling towards the more significant words. To tackle this challenge, we employ the following operation to enlarge the difference among the sampling distribution:
\begin{equation}
    \label{eq:sampling_distribution}
    P(\mathbf{I}) = \text{softmax}(\mathbf{I} * 10^{\lceil \log_{10}(\max(\mathbf{I})-\min(\mathbf{I})) \rceil}),
\end{equation}
where $\mathbf{I}=[I_{x_1},...,I_{x_L}]$. According to the probability distribution, we can obtain the replacement order $\mathbf{O}$, which is a descending order of $P(\mathbf{I})$.

\subsubsection{Updating Potential Important Words}

After determining the replacement order, we then replace each word $x_i$ with its synonym set $\mathcal{S}(x_i)$ according to the order $\mathbf{O}$. For each word, we aim to find out the best substitution, i.e., the word significantly decreases the confidence score. Thus, for each word $x_i^\prime$ in a candidate $X_t^\prime$, we have
\begin{equation}
    \label{eq:update_word}
    x_i^* = \mathop{\text{argmin}}\limits_{s \in S(x_i)} \delta(X'_{[x_i' \rightarrow s]},y, \mathbf{w}_e),
\end{equation}
where $x_i$ is the $i$-th word in $X$, $S(x_i)$ represents the synonym set of $x_i$, and $X'_{[x_i' \rightarrow s]}$ denotes $X_t'$ with the $i$-th word $x_i'$ replaced by $s$. The goal is to find the substitution that most significantly decreases the confidence score $\delta$, thereby making the prediction less confident.

Since we aim to generate transferable adversarial examples, it requires those examples that satisfy the condition $d(X'_t, X) \leq \epsilon$ listed in Eq.~\eqref{eq:definition}. However, estimating word importance using Eq.~\eqref{eq:word_importance} may not be accurate enough. To release this hard constraint, we allow the perturbation distance $d(X',X)$ to achieve a higher threshold $\eta$ that is larger than $\epsilon$ first and then remove the unimportant ones in the next step, i.e., Section~\ref{sec:remove_words}. In such a way, we can expand the scope of updated important words, effectively mitigating the impact of inaccurate word importance calculations.

In this step, we will iteratively replace each word with its best substitution obtained by Eq.~\eqref{eq:update_word} according to the order $\mathbf{O}$ until $d(X'_t, X) > \eta$. In such a way, we can obtain a potential candidate $X_t^\prime$ that is closer to achieving the desired adversarial effect.

\subsubsection{Removing Less Important Words}\label{sec:remove_words}
In the previous step, we generate an adversarial candidate $X_t^\prime$ using word substitution, but the candidate is not optimal since its distance compared with the original text $X$ exceeds the budget $\epsilon$. To address this issue, in this step, we propose to reduce the distance by putting the original word back to $X_t^\prime$.  

We then use the following steps to remove unimportant words: 
\textbf{Step 1:} For each distinct word, represented as $x_i'$ in $X'_t$, corresponding to the word $x_i$ in $X$, we substitute $x_i'$ in $X'_t$ with $x_i$, resulting in $X'_{[x_i' \rightarrow x_i]}$. We use Eq.~\eqref{eq:loss} to obtain the confidence score $\delta(X'_{[x_i' \rightarrow x_i]}, y, \mathbf{w}_e)$ again for each $X'_{[x_i' \rightarrow x_i]}$ and rank all the confidence scores in descending order. 
\textbf{Step 2:} We select the replacement word that yields the minimum score with the original word accordingly. 
\textbf{Step 3:} We check whether the new $X_t'$ satisfies the condition $d(X_t', X) < \epsilon$. If yes, then the algorithm will stop. Otherwise, we will use the new $X_t'$ to calculate the confidence score again by repeating Steps 1 - 3.

After removing the less important words, we generate a candidate $X_t'$ with the corresponding weight vector $\mathbf{w}_e$, which may be effective for the transferable adversarial attacks on the victim model. Since we have $E$ diverse weight vectors, we finally generate a set of candidates $\mathcal{X} = \{X_1', \cdots, X_{E \times T}'\}$.

\subsection{Selecting Transferable Adversarial Examples}
\label{selecting trans}

A naive solution for the transfer-based attacks is to query the victim model with each candidate $X'_t \in \mathcal{X}$. However, the number of candidates may be very large, which makes the transfer attacks inefficient. To increase the efficiency, we propose to select a set of the most transferable candidates as the adversarial examples to attack the victim model.

\begin{figure}[t]
    \centering
    \includegraphics[width=0.43\textwidth]{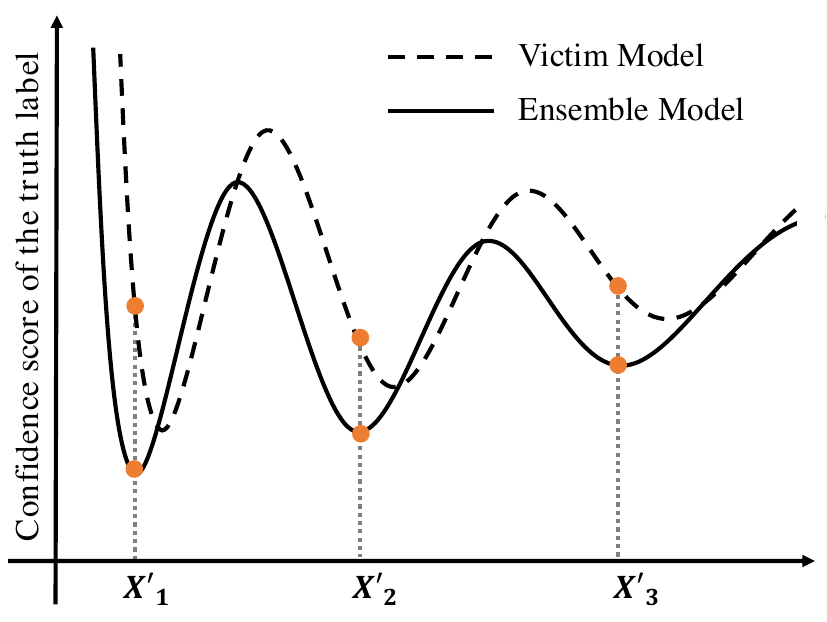}
    \caption{\small This figure illustrates the variation of confidence scores for the truth label of a sample $X'$ with respect to both the ensemble model and the victim model. In the figure, $X'_1, X'_2, X'_3$ represent potential adversarial examples.} 
    \label{fig:flat}
\end{figure}

Existing work has verified that ensemble-based adversarial attacks in images achieve good performance because the output distributions of the victim model are similar to those of the ensemble model~\cite{rap,2024HuanranChengRethinking}.
An example is shown in Figure~\ref{fig:flat}. If an adversarial example, such as $X'_1$ in Figure~\ref{fig:flat}, is located at a sharp local optimum of the ensemble model, it may perform poorly on the victim model. Similarly, if an adversarial sample is located in a flat region but with a high loss value, such as $X'_3$, its performance on the victim model may also be poor. Therefore, an ideal adversarial example with transferability should be an example with generally low loss values in its neighboring region, i.e., $X'_2$ in Figure \ref{fig:flat}.

Inspired by this, we quantify the transferability score $\tau_t$ of an example $X'_t$ by computing the average value of the examples within the adjacency region of $X'_t$. Since the text space is discrete, this adjacent region consists of samples obtained by replacing words with their synonyms. We limit the number of samples in this adjacent region based on the length of the text. 

\begin{algorithm}[t]
\caption{Selecting Transferable Adversarial Examples}
\small
\label{alg:selecting}
\SetKwInOut{Input}{Input}\SetKwInOut{Output}{Output}\SetKw{}{}
\Input{The original text $X$, the ground truth label $y$, the potential adversarial example set $\mathcal{X}$,restriction threshold $\gamma$, the ensemble weights $\bar{\mathbf{w}}$}
\Output{The examples set $\mathcal{X}'$}
\BlankLine
$m = \lceil \gamma L \rceil$ \\
\For{$X'_t \, \mathbf{in} \, \mathcal{X}$}{
    $\mathcal{C} \leftarrow $ Get the positions of all different words between $X$ and $X'_t$ \\
    $\mathcal{C}' \leftarrow$ Randomly select $m$ positions from $\mathcal{C}$\\
    \For{$i \, \mathbf{in} \, \mathcal{C}'$}{
        $s_i^* \leftarrow$ the semantically closest synonym from $S(x_i)$ \\ 
        $\beta_i = \delta(X'_{[x'_i \rightarrow s_i^*]},y;\bar{\mathbf{w}})$\\
        $\mathcal{B}_m$.insert($\beta_i$) \\
    }
    Compute $\tau_t$ by Eq. (\ref{eq:flat}); \\
    $Candidates$.insert(($X',\tau$)); \\
}
$\mathcal{X}' \leftarrow $ Select the top $K$ candidates from $Candidates$ by $\tau$ \\
\textbf{return} $\mathcal{X}'$
\end{algorithm}

Specifically, given a restriction coefficient $\gamma$ and an example $X'_t$, we define $C=\{i|x_i \neq x_i'\}$ as the set of positions where changes occur compared to $X'_t$. We randomly select $m = \lceil \gamma L \rceil$ positions from $C$. For each selected position $i$, we choose the semantically closest synonym $s_i^*$ from the synonyms set $S(x_i)$ and replace $x_i'$ with $s_i^*$ to obtain the example $X'_{[x_i' \rightarrow s_i^*]}$. This way, we can collect an adjacency region $\mathcal{B}_m(X'_t)$ consisting of $m$ examples. Based on the following formula, we can identify the transferability score $\tau_t$:
\begin{equation}
    \label{eq:flat}
    \tau_t =  - \frac{1}{|\mathcal{B}_m(X'_t)|} \mathop{\sum}\limits_{\tilde X_t \in \mathcal{B}_m(X'_t)} \delta(\tilde X_t,y,\bar{\mathbf{w}}),
\end{equation}
where $\bar{\mathbf w} = [\frac{1}{N}, \cdots, \frac{1}{N}]$ denotes the equal weights of submodels. Finally, we will rank the transferability scores in descending order and select the top $K$ candidates as the transferable adversarial examples. The detailed algorithm of selecting transferable adversarial examples is given in Algorithm \ref{alg:selecting}.

\subsection{The Overall Procedure}
The detailed algorithm procedure of SEP-Attack is given in Algorithm \ref{alg:attack}. Specifically, SEP-Attack first generates diversity-based surrogate ensemble weights. It then enters the main loop, where for each weight $\mathbf{w}_e$ , it determines the replacement order, updates potential important words, and removes less important ones. Finally, it selects the top-K candidates to attack the victim models.

\begin{algorithm}[t]
\caption{SEP-Attack}

\label{alg:attack}
\SetKwInOut{Input}{Input}\SetKwInOut{Output}{Output}\SetKw{}{}
\Input{The original text $X$, the ground truth label $y$}
\Output{The examples set $\mathcal{X}'$}
\BlankLine
Generating diversity-based surrogate ensemble weight $\mathcal{W}_{\mathcal{E}}$ \\
$X'_0 \leftarrow X$ \\
\For{$\mathbf{w}_e \, \mathbf{in} \,  \mathcal{W}_{\mathcal{E}}$}{
    \While{$t < T$} {
        Determining the replacement Order $\mathbf{O}$ \\
        \For{$i  \, \mathbf{in}  \, \mathbf{O}$}{ 
            Updating potential important words in $X'_t$\\
            \If{$d(X'_t,X)\geq \eta$}{
                break;
            }
        }
        \While{$d(X'_t,X) > \epsilon$}{
            Removing less important words of $X'_t$ \\
        }
        $\mathcal{X}$.insert($X'_t$) \\
        $t=t+1$\\
    }
}
$\mathcal{X}' \leftarrow$ selecting top $K$ candidates from $\mathcal{X}$ by Eq. \ref{eq:flat}\\
\textbf{return} $\mathcal{X}'$
\end{algorithm}

\section{Experiments}

\subsection{Experimental Settings}

\subsubsection{Dataset} \label{sec:datasets} We perform experiments on four public datasets for short and long text classification. For short text classification, we utilize the \textbf{MR} dataset~\cite{mr} and the \textbf{AG} dataset~\cite{ag_yahoo_yelp}. For long text classification, we employ the \textbf{Yelp} dataset~\cite{ag_yahoo_yelp} and the \textbf{IMDB} dataset~\cite{imdb}. Appendix \ref{sec:dataset description} provides a detailed description of these datasets.  We follow the previous method~\cite{BBA} to take 500 test examples of each dataset to conduct experiments.

\subsubsection{Surrogate and Victim Model} The models primarily employed in our experiments are BERT~\cite{BERT}, ALBERT~\cite{albert}, RoBERTa~\cite{roberta}, LSTM~\cite{LSTM}, and TextCNN~\cite{wordCNN}. Their classification accuracies are present in Table~\ref{tab:models}. The main victim models for the attack are BERT, LSTM, and CNN. When attacking a specific architecture, we take the remaining models as an ensemble. For instance, when attacking the BERT, we take ALBERT, RoBERTa, LSTM and TextCNN as the ensemble model.

\begin{table}[t]
\small
	\centering
\caption{Model performance.}
\label{tab:models}
    \begin{tabular}{l|cccc}
    \toprule
     Model       & MR(\%)  & AG(\%)   & Yelp(\%) & IMDB(\%) \\
    \midrule
    BERT    & 96.7 & 93.0  & 97.1  & 90.9  \\
    ALBERT  & 88.8 & 92.8  & 95.7  & 90.5  \\
    RoBERTa & 97.8 & 89.7  & 92.0  & 86.4  \\
    LSTM    & 83.6 & 89.9  & 94.4  & 86.7  \\
    TextCNN     & 94.6 & 90.3  & 92.5  & 85.6  \\
    \bottomrule
    \end{tabular}
\end{table}

\subsubsection{Baseline} The proposed SEP-Attack is a transfer-based attack approach based on the ensemble model. Thus, we use two \textbf{ensemble-based} methods as the key baselines. 
\begin{itemize}
    \item \textbf{Ensemble}~\cite{ensemble} is an ensemble-based method that uses TextFooler to generate adversarial examples in surrogate models and transfers them to the victim model.
    \item \textbf{HAWR}~\cite{yuan2021transferability} is also an ensemble-based method that uses word replacement rulers to generate transferable adversarial examples.
\end{itemize}

Since ensemble-based methods only query the victim model in the inference stage, we also compare them against six \textbf{query-based} textual adversarial attack methods. 
\begin{itemize}
    \item \textbf{TextFooler}~\cite{Textfooler} is a soft-label method that replaces the original word according to the word importance score.
    \item \textbf{DeepWordBug}~\cite{DeepWordBug} is an early work on soft-label attack.
    \item \textbf{TextBugger}~\cite{TextBugger} is also an early work on soft-label setting.
    \item \textbf{BBA}~\cite{BBA} is recent work on soft-label attack.
    \item \textbf{LeapAttack}~\cite{LeapAttack} is an early work on hard-label setting.
    \item \textbf{HQA-Attack}~\cite{hqa} is an recent work on hard-label setting.
\end{itemize}

\begin{table}[t]
\centering
\small
\caption{Comparison of ASR (\%) when attacking against BERT, LSTM and TextCNN models.}
\label{tab:short}
\begin{tabular}{c|l|c|ccc} 
\toprule
Dataset & Method & Budget & BERT & LSTM & TextCNN \\
\midrule
\multirow{9}{*}{MR}  & TextFooler   & \multirow{6}{*}{100}  & 47.1 & 76.1 & 64.6 \\
                     & DeepWordBug  &                       & 37.8  & 50.7 & 45.8 \\
                     & TextBugger   &                       & 45.9 & 62.0 & 54.4 \\
                     & BBA          &                       & 61.7 & 74.7 & 71.9 \\
                     & LeapAttack   &                       & 63.1  & 68.0 & 67.7 \\
                     & HQA-Attack   &                       & 61.4  & 67.5 & 70.6 \\
                     \cmidrule{2-6}
                     & Ensemble     & \multirow{3}{*}{10}   & 19.1  & 26.2 & 22.2 \\
                     & HAWR         &                       & 38.0  & 49.5 & 45.6 \\
                     & SEP-Attack &                     & \textbf{70.2}  & \textbf{78.5} & \textbf{75.9} \\
\midrule
\multirow{9}{*}{AG}  & TextFooler   & \multirow{6}{*}{100}  & 18.8 & 29.4 & 23.1 \\
                     & DeepWordBug  &    & 23.5  & 44.1 & 36.3 \\
                     & TextBugger   &    & 27.6 & 50.7 & 45.3 \\
                     & BBA          &    & 40.5 & 44.3 & 49.2 \\
                     & LeapAttack   &    & 30.4 & 33.6 & 37.6 \\
                     & HQA-Attack   &    & 33.0  & 36.2 & 36.7 \\
                     \cmidrule{2-6}
                     & Ensemble     & \multirow{3}{*}{10}   & 12.7  & 6.4 & 16.3 \\
                     & HAWR         &                      & 22.4  & 23.5 & 22.2 \\
                     & SEP-Attack &    & \textbf{58.8}  & \textbf{63.2} & \textbf{63.7} \\
\midrule
\multirow{9}{*}{Yelp}  & TextFooler   & \multirow{6}{*}{100}  & 14.8 & 29.7 & 22.5 \\
                     & DeepWordBug    &    & 27.7  & 48.8 & 48.0 \\
                     & TextBugger     &    & 45.9 & 54.4 & 62.0  \\
                     & BBA            &    & 44.0 & 61.4 & 51.3  \\
                     & LeapAttack     &    & 31.7 & 48.8 & 40.7  \\
                     & HQA-Attack     &    & 34.4 & 54.3 & 45.1  \\
                     \cmidrule{2-6}
                     & Ensemble       & \multirow{3}{*}{10}   & 27.5  & 30.0 & 27.0 \\
                     & HAWR           &                       & 42.7  & 52.8 & 52.7 \\
                     & SEP-Attack  &    & \textbf{93.0}  & \textbf{93.6} & \textbf{82.9} \\
\midrule
\multirow{9}{*}{IMDB}  & TextFooler   & \multirow{6}{*}{100}  & 7.1 & 7.4 & 10.2 \\
                     & DeepWordBug  &    & 16.5  & 29.6  & 27.3 \\
                     & TextBugger   &    & 13.6 & 16.4 & 13.4 \\
                     & BBA          &    & 45.4 & 61.6 & 56.9 \\
                     & LeapAttack   &    & 26.1 & 36.1 & 30.1 \\
                     & HQA-Attack   &    & 27.8 & 42.4 & 31.5 \\
                     \cmidrule{2-6}
                     & Ensemble     & \multirow{3}{*}{10}   & 37.4  & 46.5 & 53.0 \\
                     & HAWR         &                       & 25.9   & 65.7  & 57.7 \\
                     & SEP-Attack&   & \textbf{99.1}  & \textbf{99.5} & \textbf{98.8} \\
\bottomrule
\end{tabular}
\end{table}

\subsubsection{Evaluation Metric} We quantify the attack performance by attack success rate (\textbf{ASR}). The attack success rate is defined as the rate of successfully finding misclassified texts from the original texts that are correctly classified, which directly measures the effectiveness of the attack method. Regarding the query budget, we allocate 100 query opportunities for query-based methods and 10 query opportunities for ensemble-based methods.

\begin{figure}[t]
    \centering
    \includegraphics[width=0.35\textwidth]{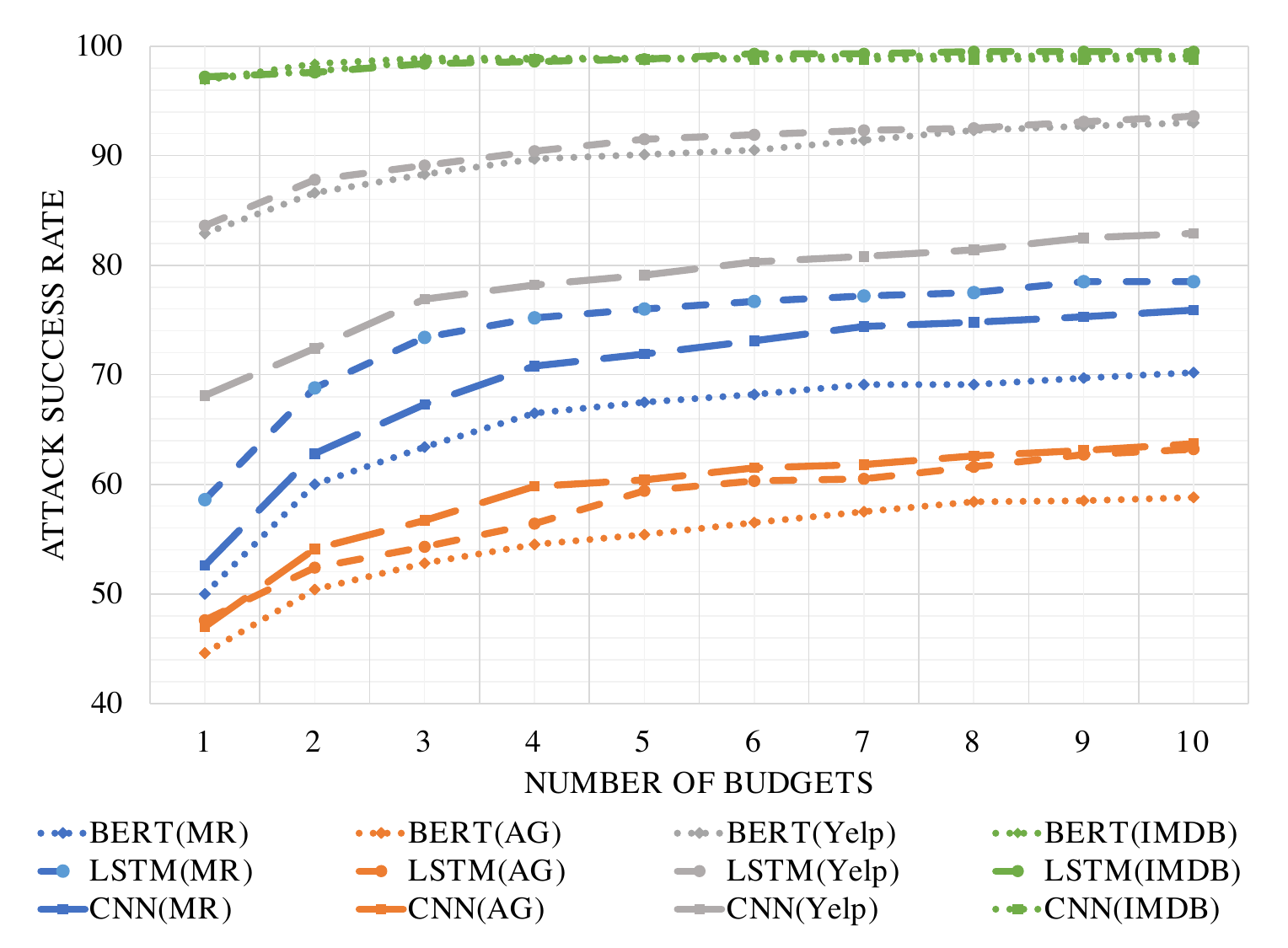}
    \caption{Comparison on ASR(\%) in different budget limits.}
    \label{fig:efficiency}
\end{figure}

\subsubsection{Implementation Details} We set $\epsilon=0.15$ in Eq. (\ref{eq:definition}), the ensemble model size $N=4$, the diversity-based surrogate ensemble weight size $E=4$, the exceed limits $\eta=0.2$ and the iteration $T=10$ for each $\mathbf{w}_e$. Additionally, we use a restriction coefficient of $\gamma=0.02$. All experiments are implemented on NVIDIA A800 GPUs. 

\begin{table}[t]
\small
\centering
\caption{Comparison of ASR (\%) when attacking against models which defend with adversarial attacks.}
\label{tab:defense}
\begin{tabular}{c|l|c|cc} 
\toprule
Victim & Method & Budget & AG & IMDB \\
\midrule
\multirow{9}{*}{HotFlip} & TextFooler   & \multirow{6}{*}{100}  & 16.0  & 7.6\\
                    & DeepWordBug  &    & 27.0        & 22.3 \\
                    & TextBugger   &    & 29.3       & 14.5 \\
                    & BBA          &    & 30.0       & 45.2 \\
                    & LeapAttack   &    & 27.6        & 23.2 \\
                    & HQA-Attack   &    & 29.1        & 24.3 \\
                    \cmidrule{2-5}
                    & Ensemble     & \multirow{3}{*}{10}   & 14.4      & 21.7 \\
                    & HAWR        &                       & 11.6          & 24.8      \\
                    &SEP-Attack         &    & \textbf{33.3}        & \textbf{85.4}\\
\midrule
\multirow{9}{*}{SHIELD} &TextFooler   & \multirow{6}{*}{100}  & 19.8 & 4.1\\
                        &DeepWordBug  &    & 20.9        & 4.75 \\
                        &TextBugger   &    & 26.4       & 4.54 \\
                        &BBA          &    & 39.2       & 37.8 \\
                        &LeapAttack   &    & 28.2        & 19.0 \\
                        &HQA-Attack   &    & 41.6        & 24.4 \\
\cmidrule{2-5}
                        & Ensemble     & \multirow{3}{*}{10}   & 7.5     & 30.5 \\
                        & HAWR        &                       & 16.1       & 23.9      \\
                        & SEP-Attack   &   & \textbf{47.1}  & \textbf{97.9}\\
\bottomrule
\end{tabular}
\end{table}

\subsection{Experimental Results}

\subsubsection{Performance Evaluation Regarding Attack Success Rates}

To demonstrate the effectiveness of our method under an extremely low query budget, we conduct attacks on BERT, LSTM, and TextCNN models along with eight baselines. The experimental results are shown in Table \ref{tab:short}. Despite our method having only one-tenth the number of queries compared to the query-based method, experimental results show that our method outperforms all other methods significantly. In short-text datasets like the AG dataset, when attacking BERT, LSTM, and TextCNN, our method achieves attack success rate improvements of 18.3\%, 12.5\%, and 14.5\%, respectively, compared to the second-ranked method. In long-text datasets, the performance of our method is even more remarkable. For example, on the IMDB dataset, when attacking BERT, LSTM, and TextCNN, our attack success rates are close to 100\%, with improvements of 53.7\%, 33.8\%, and 41.1\%, respectively, compared to the second-ranked method. This indicates that our method achieves high attack success rates with a low query budget, demonstrating the effectiveness of our approach.

Figure \ref{fig:efficiency} illustrates the performance of our method attacking different models on diverse datasets across varying query budgets. The figure shows that our proposed SEP-Attack achieves significant effectiveness even at a query budget of 5, providing further evidences of the efficacy of our method.

\subsubsection{Attack Models with Safety Guard Mechanisms}

With the development of adversarial attacks, AI security issues gain increasing attention, leading to the emergence of defense methods against adversarial attacks. In order to demonstrate the effectiveness of our attack method in bypassing defense methods, we conduct experiments on attacking defense models. We use HotFlip \cite{hotflip} to generate adversarial examples for adversarial training on BERT as the first victim model. As the second victim model, we employ the adversarial defense method form SHIELD \cite{SHIELD} with BERT as the architecture. We fine-tune these two models on the short text dataset AG and the long text dataset IMDB respectively. HotFlip achieves a refined classification success rate of 91.6\% on the AG dataset and 89.6\% on the IMDB dataset after fine-tuning BERT. Similarly, Shield achieves a refined classification success rate of 88.0\% on both the AG and IMDB datasets after fine-tuning BERT. 

The experimental results are presented in Table \ref{tab:defense}. Regardless of whether HotFlip or SHIELD is used as the victim model, the attack success rates of SEP-Attack significantly surpass that of previous methods. In the case of HotFlip, we achieve attack success rate improvements of 3.3\%, and 40.2\% on the AG and IMDB datasets, respectively. For SHIELD, we observe even greater attack success rate improvements on the AG and IMDB datasets, reaching 5.5\% and 60.1\%, respectively. Notably, when attacking SHIELD on the IMDB dataset, our attack success rate is 97.9\%, achieving a near-perfect rate. These experimental results demonstrate the capability of our method to bypass the limitations of defense models, thereby inspiring the search for more effective adversarial defense methods.

\subsubsection{Attack Real-World APIs}

Sentiment analysis has widespread applications in the real world. To evaluate the effectiveness of our method in attacking real-world APIs, we conduct experiments targeting Alibaba Cloud and Google Cloud. Due to the limited query volume on both platforms, we compare SEP-Attack with BBA, LeapAttack, HQA-Attack, Ensemble, and HAWR. We intergrat ALBERT, RoBERTa, LSTM, and TextCNN as the ensemble model and randomly select 100 data points from Yelp for experiments. The results are summarized in Table \ref{tab:real_world}. The results clearly indicate that SEP-Attack achieves higher attack success rates than other methods, making it more proficient in attacking real-world APIs.

\begin{table}[t]
\small
\caption{Comparison of ASR(\%) when attacking against real-world APIs.}
        \label{tab:real_world}
        \centering
        \begin{tabular}{c|l|c|cc} 
        \toprule
        Dataset & Method & Budget & Alibaba & Google \\
        \midrule
        \multirow{7}{*}{Yelp}
                            & TextBugger   & \multirow{3}{*}{100} & 56.5  & 17.5           \\
                            & BBA          &    & 58.7       & 30.9 \\
                            & LeapAttack   &    & 59.8       & 29.9                     \\
                            & HQA-Attack   &    & 55.4       & 37.1                     \\
                            \cmidrule{2-5}
                            & Ensemble     & \multirow{3}{*}{10} & 18.6      & 16.5       \\
                            & HAWR        &                     & 9.8      & 18.6       \\
                            &SEP-Attack &    & \textbf{64.1}        & \textbf{75.3}\\
        \bottomrule
        \end{tabular}
\end{table}
\vspace{-2pt}

\subsubsection{Attack Large Language Models}

With the increasing size and enhanced performance of large language models (LLMs), their susceptibility to adversarial attacks has emerged as a crucial research focus. To evaluate the performance of our model on large language models, we conduct an ensemble-based attack, similar to real-world API attacks, on Meta-Llama-3-8B-Instruct \cite{llama3modelcard}, Mistral-7B-Instruct-v0.2 \cite{mistral7b}, and GPT-3.5, using the AG and Yelp datasets to assess their robustness. The classification accuracies of the models are as follows: Llama achieves 81.6\% accuracy on AG and 94.2\% on Yelp; Mistral achieves 75.0\% accuracy on AG and 93.6\% on Yelp; and GPT-3.5 achieves 82.4\% accuracy on AG and 95.0\% on Yelp. The experimental results of the attack are presented in Table \ref{tab:llm6}, demonstrating that our proposed SEP-Attack method achieves significantly higher attack success rates (ASR) compared to the other two attack methods across all the tested models and datasets. For example, on the Llama model, SEP-Attack achieves an ASR of 38.7\% on AG and 65.0\% on Yelp, which are much higher than the ASR of Ensemble and HAWR. Similar improvements are observed on Mistral and GPT-3.5, demonstrating the superior capability of SEP-Attack in exploiting LLM vulnerabilities.

\begin{table}[t]
\small
\caption{Comparison of ASR(\%) when attacking against large language models.}
\label{tab:llm6}
\begin{tabular}{c|l|cc}
\toprule
Victim & Method & AG & Yelp \\
\midrule
\multirow{3}{*}{Llama} & Ensemble & 10.3 & 9.1 \\
                       & HAWR     & 18.8 & 12.2 \\
                       & SEP-Attack & \textbf{38.7} & \textbf{65.0} \\
\midrule
\multirow{3}{*}{Mistral} & Ensemble & 16.5 & 16.8 \\
                         & HAWR     & 20.4 & 22.3 \\
                         & SEP-Attack & \textbf{41.6} & \textbf{72.9} \\
\midrule
\multirow{3}{*}{ChatGPT} & Ensemble & 12.8 & 16.3 \\
                         & HAWR     & 15.6 & 14.5 \\
                         & SEP-Attack & \textbf{32.3} & \textbf{64.4} \\
\bottomrule
\end{tabular}
\end{table}

\subsubsection{Parameter Investigation}
In our parameter investigation, we conduct experiments with a query budget of 10 to evaluate the performance of SEP-Attack under low-query-budget constraints. We explore the impact of two key parameters, $\eta$ and $\gamma$, on the attack success rate. The detailed results are summarized in Tables \ref{tab:eta} and \ref{tab:gamma}. For the perturbation budget $\eta$, we find that higher values generally led to better attack success rates but at the cost of increased runtime. Specifically, $\eta=0.200$ provide a good balance between success rate and efficiency. For the parameter $\gamma$, our experiments show that $\gamma=0.02$ achieves better performance than $\gamma=0$, demonstrating the effectiveness of the candidate selection step described in Section \ref{selecting trans}. These findings highlight the importance of carefully tuning these parameters to optimize the performance of SEP-Attack under limited query budgets.

\begin{table}[t]
\small
    \centering
        \caption{The adversarial attack performance of SEP-Attack with different $\eta$ when attacking BERT on AG and Yelp.}
    \label{tab:eta}
    \begin{tabular}{c|cccccc}
    \toprule
        \multirow{2}{*}{Dataset} & \multicolumn{2}{c}{0.175} &\multicolumn{2}{c}{0.200}  &\multicolumn{2}{c}{0.225} \\
        \cmidrule{2-7}
          & ASR & Time(s) & ASR & Time(s) & ASR & Time(s)  \\
        \midrule
        AG & 50.2 & 16 & 58.8 & 20.6 & 59.6 & 30 \\
        Yelp & 86.7 & 207 & 93.0 & 245.4 & 94.2 & 289 \\
        \bottomrule
    \end{tabular}
\end{table}

\begin{table}[t]
\small
    \centering
    \caption{The adversarial attack performance of SEP-Attack with different $\gamma$ when attacking BERT on AG and Yelp.}
    \label{tab:gamma}
    \begin{tabular}{c|cccccccc}
    \toprule
        Dataset & 0 & 0.01 & 0.02 & 0.03 & 0.05 & 0.1 & 0.2 & 0.3\\
        \midrule
        AG   & 58.4 & 58.6 & 58.8 & 58.8 & 58.2 & 58.6 & 58.4 & 58.4 \\
        Yelp & 93.0 & 92.8 & 93.0 & 92.8 & 92.8 & 93.0 & 92.8 & 92.7 \\
        \bottomrule
    \end{tabular}
\end{table}

\subsubsection{Ablation Study and Case Study}

In order to demonstrate the effectiveness of each module, we perform ablation studies on the three datasets when attacking BERT. The results are shown in Table \ref{tab:ablation}. Random Weights refers to the surrogate ensemble weights generated through random sampling. w/o Selecting means that the SEP-Attack without the selecting transferable adversarial examples step. The experimental findings highlight the importance of both the diversity-based weights and the selection of transferable adversarial examples in achieving improved performance. We also list some concrete adversarial examples generated by SEP-Attack, which are shown in Appendix \ref{sec:case_study}.

\begin{table}[t]
\small
    \centering
    \caption{The ablation study of SEP-Attack when attacking against BERT on different datasets.}
    \label{tab:ablation}
    \begin{tabular}{c|ccc}
    \toprule
        Dataset   &  Random Weigths & w/o Selecting & SEP-Attack\\
    \midrule
        AG & 57.3 & 58.4 & \textbf{58.8} \\
        Yelp & 89.7 & 92.7 & \textbf{99.1} \\
    \bottomrule
    \end{tabular}
\end{table}

\section{Conclusion}
In this paper, we proposed SEP-Attack, a simple yet effective paradigm for transfer-based textual adversarial attacks, addressing the challenge of transferability in the text domain. Experiments on four datasets and two real-world APIs show that SEP-Attack consistently surpasses state-of-the-art baselines, demonstrating its robustness and practical value. Beyond improving attack performance, this work highlights the need to consider submodel diversity and reliable importance estimation when evaluating the security of Web-facing language models. Our findings provide insights that support building safer and more trustworthy Web AI systems. In future work, we plan to explore the theoretical foundations of SEP-Attack and extend it to other modalities.

\section{Acknowledgements}
This work was supported by National Natural Science Foundation of China (No. 62206038, 62106035), the Strategic Priority Research Program of the Chinese Academy of Sciences (No. XDA0490301), Liaoning Binhai Laboratory Project (No. LBLF-2023-01), Xiaomi Young Talents Program, and the Interdisciplinary Institute of Smart Molecular Engineering, Dalian University of Technology.

\bibliographystyle{ACM-Reference-Format}
\bibliography{sample-base}

\appendix

\section{Dataset Description}
\label{sec:dataset description}
The detailed dataset information is as follows:
\begin{itemize}
    \item \textbf{MR} \cite{mr} is a short movie review dataset for binary sentiment classification.
    \item \textbf{AG} \cite{ag_yahoo_yelp} is a news topic classification dataset with four categories.
    \item \textbf{Yelp} \cite{ag_yahoo_yelp} is a binary sentiment classification dataset.
    \item  \textbf{IMDB} \cite{imdb} is a movie review dataset for binary sentiment, with longer reviews compared to MR.
\end{itemize}

\section{Case Study}
\label{sec:case_study}

We list some adversarial examples generated by SEP-Attack on MR, AG, Yelp and IMDB datasets which are in Table \ref{table:case_study}. Take the first adversarial example in Table \ref{table:case_study} as an example, just replacing the word ``wonderful'' with ``extraordinaire'' can change the prediction from “positive” to “negative”, which demonstrates that SEP-Attack can generate an effective adversarial example.

\begin{table}[b]
\centering
\caption{Adversarial examples generated by SEP-Attack in different datasets against the BERT model. The substituted original word is in the italics format, and the replacement is the following one.}
\label{table:case_study}
\begin{tabular}{l|c} 
\toprule
\multicolumn{1}{c|}{Adversarial Example} & Change of prediction \\ 
\midrule
\begin{tabular}[c]{@{}l@{}}\textbf{MR:} it's weird, \textit{wonderful} \\ (\textbf{extraordinaire}), and not necessarily  \\ for kids. \end{tabular}& Positive $\rightarrow$ Negative   \\ 
\midrule
\begin{tabular}[c]{@{}l@{}}\textbf{AG:} unskilled \textit{jobs} (\textbf{tasks}) to go there \\
will be no jobs for unskilled \\
\textit{workers} (\textbf{assistants}) in britain within\\
10 years, the leading \\
\textit{employers} (\textbf{employing}) \# 39;\\
\textit{organisation} (\textbf{organisms}) claims today.\\
the prediction is based on the growth in \end{tabular}&  Bussiness$\rightarrow$ Sci/Tech   \\ 
\midrule
\begin{tabular}[c]{@{}l@{}}\textbf{Yelp:}  
tram has \textit{great} (\textbf{whopping}) pho , \\
and enough of it to feed all of \\
pittsburgh when the \textit{next} (\textbf{follows}) \textit{fake} \\
(\textbf{fraudulent}) armageddon rolls around.
\end{tabular}& Positive $\rightarrow$ Negative   \\ 
\midrule
\begin{tabular}[c]{@{}l@{}}\textbf{IMDB:}
focus is an \textit{engaging} (\textbf{implicate}) \\
story told in urban , wwii era setting \\
william macy \textit{portrays} (\textbf{exemplifies}) \\
everyman who is taken out of his \\
personal \textit{circumstances} (\textbf{screenplays}) \\
and \textit{challenged} (\textbf{disproved}) with \\
decisions testing his values affecting \\
the community laura dern, macy and \\
david paymer give \textit{good} (\textbf{adequate}) \\
performances, so \textit{also} (\textbf{instead}) \\
the good \textit{supporting} (\textbf{aid}) ensemble \end{tabular}  & Positive $\rightarrow$ Negative\\
\bottomrule
\end{tabular}
\end{table}

\end{document}